# Application of Rough Set Theory to Analysis of Hydrocyclone Operation

H.Owladeghaffari, M.Ejtemaei&M.Irannajad

*Department of Mining&Metulrgical Engineering, Amirkabir University of Technology, Tehran, Iran*

**Outline**: This paper describes application of rough set theory, on the analysis of hydrocyclone operation. In this manner, using Self Organizing Map (SOM) as preprocessing step, best crisp granules of data are obtained. Then, using a combining of SOM and rough set theory (RST)-called SORST-, the dominant rules on the information table, obtained from laboratory tests, are extracted. Based on these rules, an approximate estimation on decision attribute is fulfilled. Finally, a brief comparison of this method with the SOM-NFIS system (briefly SONFIS) is highlighted.

**A brief introuduction:**

In the design of mineral processing cycle, one of the most important issues is the selection of hydrocyclone in different parts of the site. However, prediction of hydrocyclone performance using direct or indirect modeling faces with some difficulties. Apart from analytical, numerical, or experimental modeling, modeling based on intelligent systems can be supposed as an excellent situation, which is ensued by data engineering, machine learning, and stochastic learning theorems.

With advancing and extension of intelligent knowledge discovery (Data mining), in different applied sciences, selection of best features, accounting of the uncetainty in the monitored data, are the main challenges of the most sciences.

Because of being the uncertainty feature of the monitored data, accounting of uncertainty approaches such probability, fuzzy set and rough set theories to knowledge acquisition, extraction of rules and prediction of unknown cases, have been distinguished, more than the past. The granulation of information theory [1] covers the mentioned approaches in two formats: crisp information granulation and fuzzy information granulation.

There are two reasons why we propose this concept to tackle uncertainty in the monitored mineral processing data. The first one is human instinct. As human beings, we have developed a granular view of the world. When describing a problem, we tend to shy away from numbers and use aggregates to ponder the question instead. This is especially true when a problem involves incomplete, uncertain, or vague information. It may be sometimes difficult to differentiate distinct elements, and so one is force consider "information granules (IG) which one collection of entities arranged together due to their similarity, functional adjacency, and indistinguishability.

In this study, using three computational intelligence (CI) theories, neural networks, fuzzy inference system and rough set based on information granulation theory, two algorithms to analyses of hydrocyclone data will be presented. In our models, self-organizing feature map, Neuro-Fuzzy Inference System and rough set are utilized to construct IGs. Details of our instruments can be followed in [2], [3], [4].

**Proposed algorithms:**

In the whole of our algorithms, we use four basic axioms upon the balancing of the successive granules:
Step (1): dividing the monitored data into groups of training and testing data
Step (2): first granulation (crisp) by SOM or other crisp granulation methods
  Step (2-1): selecting the level of granularity randomly or depend on the obtained error from the NFIS or RST (regular neuron growth)
  Step (2-2): construction of the granules (crisp).
Step (3): second granulation (fuzzy or rough IGs) by NFIS or RST
  Step (3-1): crisp granules as a new data.
  Step (3-2): selecting the level of granularity; (Error level, number of rules, strength threshold...)
  Step (3-3): checking the suitability. (Close-open iteration: referring to the real data and reinspect closed world)
  Step (3-4): construction of fuzzy/rough granules.
Step (4): extraction of knowledge rules
  Selection of initial crisp granules can be supposed as "Close World Assumption (CWA)" .But in many applications, the assumption of complete information is not feasible, and only cannot be used. In such cases, an "Open World Assumption (OWA)', where information not known by an agent is assumed to be unknown, is often accepted [5].
Balancing assumption is satisfied by the close-open iterations: this process is a guideline to balancing of crisp and sub fuzzy/rough granules by some random/regular selection of initial granules or other optimal structures and increment of supporting rules (fuzzy partitions or increasing of lower /upper approximations ), gradually.
The overall schematic of Self Organizing Neuro-Fuzzy Inference System -Random and Regular neuron growth-: SONFIS-R, SONFIS-AR; has been shown in figure1.
In first regular granulation, we use a linear relation is given by:
$$N_{t+1} = \alpha N_t + \Delta_t ; \Delta_t = \beta E_t + \gamma$$
Where $N_t = n_1 \times n_2; |n_1 - n_2| = Min.$ is number of neurons in SOM; $E_t$ is the obtained error (measured error) from second granulation on the test data and coefficients must be determined, depend on the used data set. Obviously, one can employ like manipulation in the rule (second granulation) generation part, i.e., number of rules.
Determination of granulation level is controlled with three main parameters: range of neuron growth, number of rules and error level. The main benefit of this algorithm is to looking for best structure and rules for two known intelligent system, while in independent situations each of them has some appropriate problems such: finding of spurious patterns for the large data sets, extra-time training of NFIS or SOM.
In second algorithm, apart from employing hard computing methods (hard granules), RST instead of NFIS has been proposed (figure 2). Applying of SOM as a preprocessing step and discretization tool is second process. Categorization of attributes (inputs/outputs) is transferring of the attribute space to the symbolic appropriate attributes.
In fact for continuous valued attributes, the feature space needs to be discretized for defining indiscernibilty relations and equivalence classes. We discretize each feature in to some levels by SOM, for example "low, medium, and high" for attribute "a". Finer discretization may lead to better accuracy to recognizing of test data but imposes the higher cost of a computational load.
Because of the generated rules by a rough set are coarse and therefore need to be fine-tuned, here, we have used the preprocessing step on data set to crisp granulation by SOM (close world assumption).

In fact, with referring to the instinct of the human, we understand that human being want to states the events in the best simple words, sentences, rules, functions and so forth. Undoubtedly, such granules while satisfies the mentioned axiom that describe the distinguished initial structure(s) of events or immature data sets. Second SOM, as well as close world assumption, gets such dominant structures on the real data. In other word, condensation of real world and concentration on this space is associated with approximate analysis, such rough or fuzzy facets.

**Results:**

Analysis of first situation is started off by setting number of close-open iteration and maximum number of rules equal to 10 and 4 in SONFIS-R, respectively. The error measure criteria in SONFIS are Root Mean Square Error (RMSE), given as below:

$$RMSE = \sqrt{\frac{\sum_{i=1}^{m}(t_i - t_i^*)^2}{m}} \;;$$

Where $t_i$ is output of SONFIS and $t_i^*$ is real answer; m is the number of test data (test objects). In the rest of paper, let m=19 and number of training data set =150. Figures 3 indicate the results of the aforesaid system. The indicated position in figure 3a, b states minimum RMSE over the iterations.

It is worth noting that upon this balancing criterion, we may loose the general dominant distribution on the data space. The performance of the obtined fyzzy rules on the test data has been portrayed in figure 4(a). So, the membership functions of each input can be compared by thr real training distribution (figure 4b).

In employing of second algorithm (figure2), we use- -for in this case- only exact rules i.e., one decision class in right hand of an if-then rule. Figure 14 and 15 depict the scaling process by 1-D SOM (3 neurons) and the performance of SORST-R over 7 random selection of SOM structure, respectively. The applied Error measure is :

$$EM = \frac{\sum_{i=1}^{m}(d_i^{real} - d_i^{classified})^2}{m} \;;$$

It must be noticed that for unrecognizable objects in test data (elicited by rules) a fix value such 4 is ascribed. So for measure part when any object is not identified, 1 is attributed. This is main reason of such swing of EM in reduced data set 6 (figure 5-b). Clearly, in data set 5 SORST gains a lowest error (15 neurons in SOM).

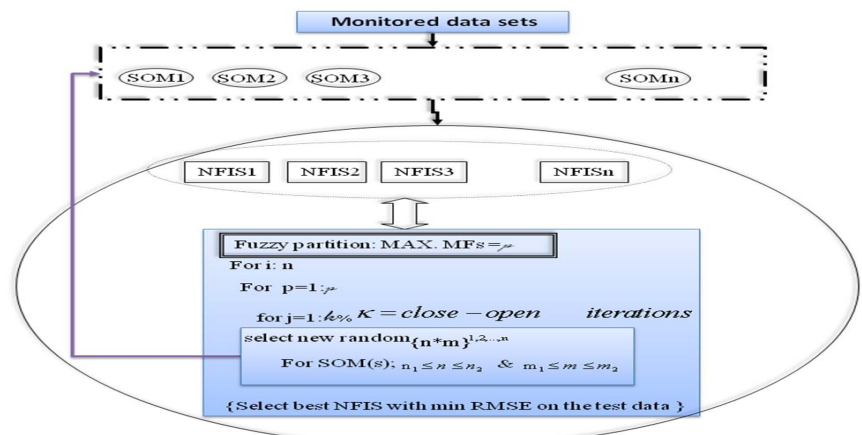

Figure 1. Self Organizing Neuro-Fuzzy Inference System (SONFIS)

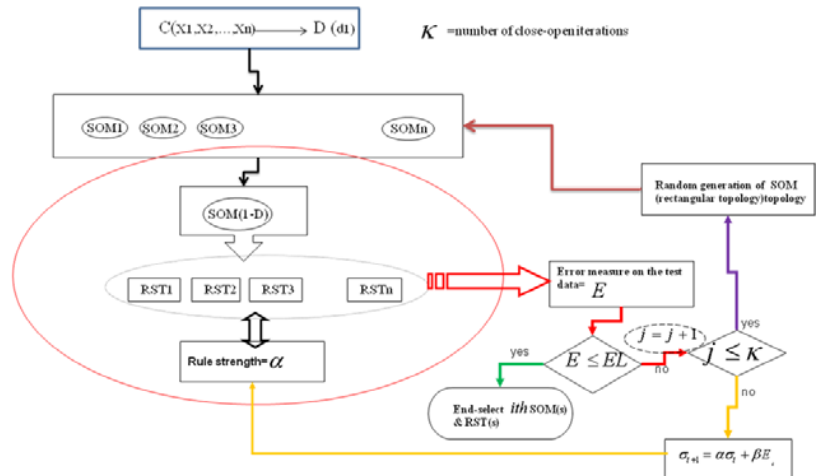

Figure 2. Self Organizing Rough Set Theory-Random neuron growth & adaptive strength factor (SORST-R)

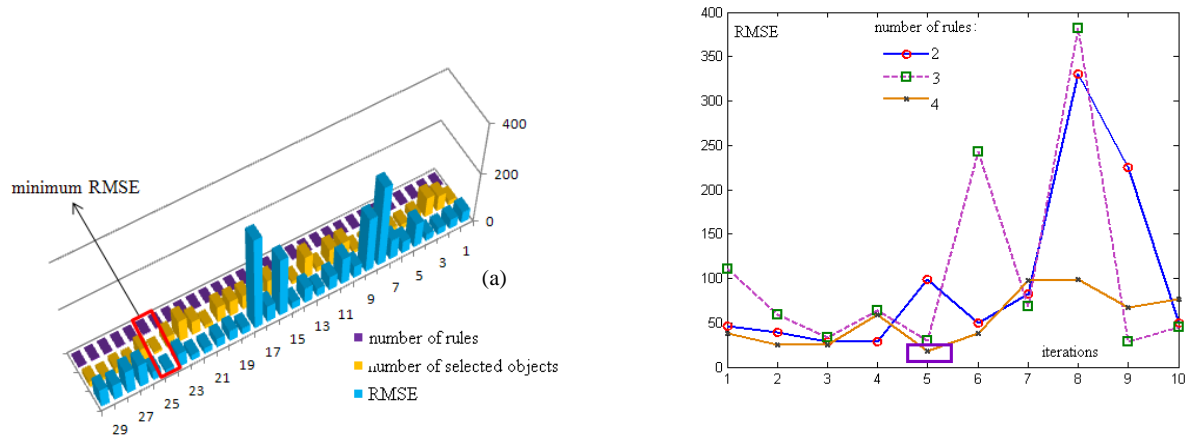

Figure3. obtanined results by SONFIS-R and the minimum RMSE in 30 iteration -10 for each rule

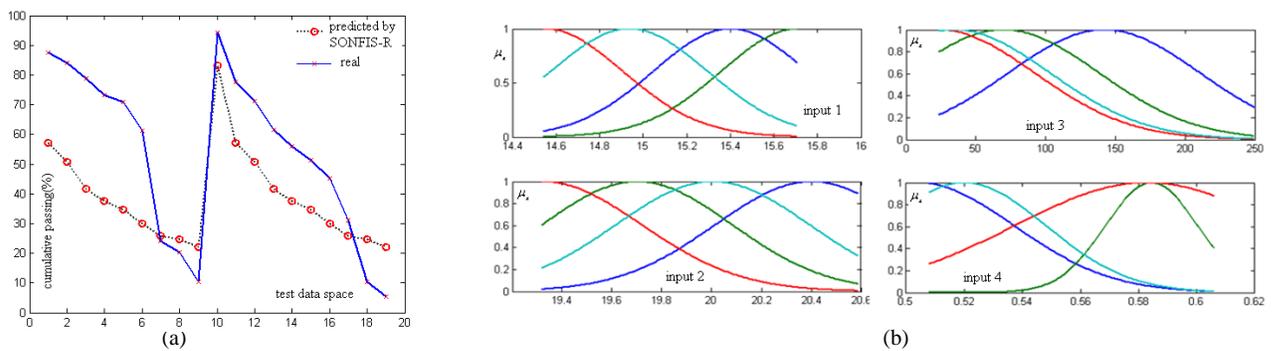

Figure4.a)the real and predicted decision on the testing data set with sub-fuzzy granulation; b) fuzzy granulation of inputs ;vertical axises are memebership degree( $\mu_x$ )of any input.

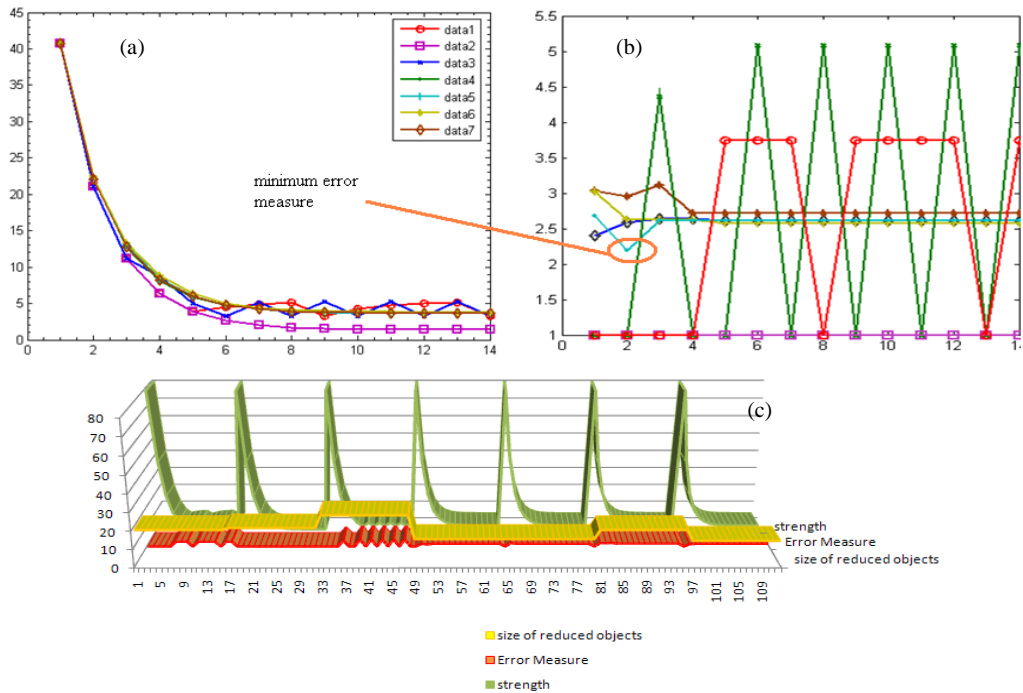

Figure 5. SORST-R results on the Hydrocyclon data set: a) strength factor convergence (approximately); b) error measure variations along strength factor updating and c) 3-D column perspective of error measure- size of reduced objects